\documentclass[twocolumn]{el-author}

\newcommand{\ua}{\uparrow}
\newcommand{\nc}{\newcommand}
\nc{\da}{\downarrow} \nc{\hc}{\hat{c}} \nc{\hS}{\hat{S}}
\nc{\bra}{\langle} \nc{\ket}{\rangle} \nc{\eq}{equation (\ref}
\nc{\h}{\hat} \nc{\hT}{\h{T}}\nc{\be}{\begin{eqnarray}}
\nc{\ee}{\end{eqnarray}}\nc{\rd}{\textrm{d}}\nc{\e}{eqnarray}\nc{\hR}{\hat{R}}\nc{\Tr}{\mathrm{Tr}}
\nc{\tS}{\tilde{S}}\nc{\tr}{\mathrm{tr}}\nc{\8}{\infty}\nc{\lgs}{\bra\ua,\phi|}\nc{\rgs}{|\ua,\phi\ket}
\nc{\hU}{\hat{U}}\nc{\lfs}{\bra\phi|}\nc{\rfs}{|\phi\ket}\nc{\hZ}{\hat{Z}}\nc{\hd}{\hat{d}}\nc{\mD}{\mathcal{D}}
\nc{\bd}{\bar{d}}\nc{\bc}{\bar{c}}\nc{\mc}{\mathcal}\nc{\ea}{eqnarray}\nc{\mG}{\mathcal{G}}\nc{\bce}{\begin{center}}
\nc{\ece}{\end{center}}
\date{12th December 2011}

\usepackage{amssymb,xspace,intmacros,subfigure,amsfonts}

\newcommand{\wb}{{\textbf{w}}}
\newcommand{\xb}{{\textbf{x}}}

\newcommand{\lone}{$\ell^{1}$}

\newcommand{\yb}{{\textbf{y}}}
\newcommand{\sbb}{{\textbf{s}}}
\newcommand{\ab}{{\textbf{a}}}

\newcommand{\zb}{{\textbf{z}}}

\newcommand{\Eb}{{\textbf{E}}}
\newcommand{\Ab}{{\textbf{A}}}
\newcommand{\nb}{{\textbf{n}}}
\newcommand{\Ib}{{\textbf{I}}}
\newcommand{\qb}{{\textbf{q}}}
\newcommand{\rb}{{\textbf{r}}}
\newcommand{\hb}{{\textbf{h}}}
\newcommand{\vb}{{\textbf{v}}}

\newcommand\ie{i.e.\xspace}

\begin{document}

\title{Bayesian hypothesis testing for one bit compressed sensing with sensing matrix perturbation}

\author{H. Zayyani, M. Korki and F. Marvasti}

\abstract{This letter proposes a low-computational Bayesian algorithm for noisy sparse recovery in the context of one bit compressed sensing with sensing matrix perturbation. The proposed algorithm which is called BHT-MLE comprises a sparse support detector and an amplitude estimator. The support detector utilizes Bayesian hypothesis test, while the amplitude estimator uses an ML estimator which is obtained by solving a convex optimization problem. Simulation results show that BHT-MLE algorithm offers more reconstruction accuracy than that of an ML estimator (MLE) at a low computational cost.}

\maketitle

\section{Introduction}
The one bit compressed sensing which is the extreme case of quantized compressed sensing \cite{ZymnBC10} has been extensively investigated recently [2-9]. 
In the one bit compressed sensing framework, it is proved that accurate and stable recovery can be achieved by using only the sign of linear measurements \cite{JacqLBB13}.

Many algorithms have been developed for one bit compressed sensing. A renormalized fixed-point iteration (RFPI) algorithm which is based on \lone-norm minimization has been presented in \cite{BoufB08}. Also, a matching sign pursuit (MSP) algorithm has been proposed in \cite{Bouf09}. A binary iterative hard thresholding (BIHT) algorithm introduced in \cite{JacqLBB13}, has been shown to have better performance than that of MSP. Moreover, a restricted-step shrinkage (RSS) algorithm which has been devised in \cite{Lask11} has provable convergence guarantees.

 In addition to noise-free settings, there may be noisy sign measurements. In the noisy case, the sign flips may worsen the performance. In \cite{YanYO12}, an adaptive outlier pursuit (AOP) algorithm is developed to detect the sign flips and reconstruct the signals with very high accuracy even when there are a large number of sign flips \cite{YanYO12}. Moreover, noise-adaptive RFPI (NARFPI) algorithm combines the idea of RFPI and AOP \cite{MovaPD12}. Recently, a one bit Bayesian compressed sensing \cite{Li15} and a MAP approach \cite{DongZ15} have been developed for solving the problem. Also, \cite{Zhu14} focuses on the ML estimation of a vector parameter from sign measurements with sensing matrix perturbation.

 In this letter, similar to some sparse recovery algorithms (see e.g., \cite{ZayyBJ09}), we propose a two-step approach for one bit compressed sensing with sensing matrix perturbation. The first step is the support detection and the second step is the amplitude estimation. In the first step, a Bayesian hypothesis test is used to detect the active samples of sparse vector. On the other hand, for amplitude recovery of active samples of the sparse vector, similar to \cite{Zhu14}, we utilize an MLE. Compared to \cite{Zhu14} which ignores the sparsity, the main advantage of our proposed algorithm is to exploit the sparsity of the sparse vector. Moreover, detecting the active samples in the first step, reduces the complexity of the optimization problem of MLE. Our simulation results verify that exploiting the sparsity enhances the reconstruction performance of the sparse vector.


\section{Problem Formulation}
Consider a sparse vector $\sbb$ which is observed via a corrupted sensing matrix as
\begin{equation}
\xb=(\Ab+\Eb)^T\sbb+\nb,
\end{equation}
where $\Ab \in \mathbb{R}^{m\times N}$ is a known sensing matrix, $\Eb$ is an error random matrix whose elements are i.i.d with $e_{ij}\sim\mathcal{N}(0,\sigma^2_e)$ where $\sigma^2_e$ is viewed as perturbation strength, and $\nb$ is the additive noise vector (independent of $\Eb$) with $\nb\sim\mathcal{N}(0,\sigma^2_n\Ib)$.

In one bit compressed sensing with sensing matrix perturbation, we aim to estimate the sparse vector $\sbb$ based on the sign of linear measurements $\yb=\mathrm{sign}(\xb)$ which is
\begin{equation}
\yb=\mathrm{sign}(\Ab^T\sbb+\zb),
\end{equation}
where $\zb=\Eb^T\sbb+\nb$ is  called equivalent noise which is the sum of a multiplicative noise and an additive noise \cite{Zhu14}. It can be simply shown that the variance of the noise $\zb\sim\mathcal{N}(0,\sigma^2_z\Ib)$ is \cite{Zhu14}
\begin{equation}
\sigma^2_z=||\sbb||^2_2\sigma^2_e+\sigma^2_n.
\end{equation}

The sparse vector $\sbb$ is assumed to have a Bernoulli-Gaussian distribution, \ie $s_j=q_jr_j$, where $q_j$ is the activity of the $j$'th element of sparse vector and $r_j$ is the amplitude of the element. Similar to some sparse recovery algorithms \cite{ZayyBJ09}, the sparse recovery is equivalent to estimating both the activity vector $\qb=[q_1,q_2,...,q_m]^T$ and amplitude vector $\rb=[r_1,r_2,...,r_m]$.

The proposed algorithm is divided into two steps. In the first step, we estimate the activity vector $\qb$. We call this support detection. We call the second step of our proposed algorithm amplitude recovery. In this step, we estimate the amplitude vector  $\rb$.

\section{Support Detection Using Bayesian Hypothesis Testing}
For determining the activity of $j$'th element of sparse vector $\sbb$, we consider two hypotheses. The first hypothesis $H_{1j}$ assumes that $s_j$ is inactive and the second hypothesis $H_{2j}$ considers that $s_j$ is active. We have
\begin{equation}
H_{1j}:\Bigg\{\begin{array}{cc}
                    y_1=\mathrm{sign}(\ab^T_1\sbb_{-j}+z_1), &  \\
                    y_2=\mathrm{sign}(\ab^T_2\sbb_{-j}+z_2), &  \\
                    ... &  \\
                    y_N=\mathrm{sign}(\ab^T_N\sbb_{-j}+z_N), & \qquad
                  \end{array}
\end{equation}
and
\begin{equation}
H_{2j}:\Bigg\{\begin{array}{cc}
                    y_1=\mathrm{sign}(\ab^T_1\sbb+z_1), &  \\
                    y_2=\mathrm{sign}(\ab^T_2\sbb+z_2), &  \\
                    ... &  \\
                    y_N=\mathrm{sign}(\ab^T_N\sbb+z_N), & \qquad
                  \end{array}
\end{equation}
where $\sbb_{-j}$ is the sparse vector $\sbb$ with $s_j=0$, $\ab_i^T$ is the $i$'th measurement vector, $y_i$ is the $i$'th sign measurement element and $N$ is the number of measurements. The Bayesian hypothesis test is
\begin{equation}
\hat{q}_j=\Big\{\begin{array}{cc}
                    0 & p(H_{1j}|\yb)\ge p(H_{2j}|\yb), \\
                    1 & Othewise,\qquad
                  \end{array}
\end{equation}
where $\yb=[y_1,y_2,...,y_N]^T$ is the sign measurement vector. Using the MAP rule, the activity rule based on the hypothesis test is $p(H_{1j})p(\yb|H_{1j})\le p(H_{2j})p(\yb|H_{2j})$.
Assume that the prior probabilities are $p(H_{1j})=1-p$ and $p(H_{2j})=p$, where $p$ is the activity probability. Also, assuming the independence of measurements $y_i$, we have $p(\yb|H_{1j})=\prod_{i=1}^N p(y_i|H_{1j})$ and $p(\yb|H_{2j})=\prod_{i=1}^N p(y_i|H_{2j})$. Taking the logarithm with some manipulations leads to the following activity rule:
\begin{equation}
\sum_{i=1}^N\ln(p(y_i|H_{2j}))-\sum_{i=1}^N\ln(p(y_i|H_{1j}))\ge\mathrm{Th}=\ln(\frac{1-p}{p}).
\end{equation}

Simple calculations show that $p(y_i|H_{1j})=\Phi(\frac{y_i\ab^T_i\sbb_{-j}}{\sigma_z})$, where $\Phi(u)=\frac{1}{\sqrt{2\pi}}\int_{-\infty}^u e^{-\frac{x^2}{2}}dx$ is the cumulative distribution function of the standard Gaussian distribution. Similarly, we have $p(y_i|H_{2j})=\Phi(\frac{y_i\ab^T_i\sbb}{\sigma_z})$. Hence, the overall hypothesis test for determining the activity of $j$'th element is
\begin{equation}
\label{eq: act}
\sum_{i=1}^N\ln(\frac{\Phi(\frac{y_i\ab^T_i\sbb}{\sigma_z})}{\Phi(\frac{y_i\ab^T_i\sbb_{-j}}{\sigma_z})})\ge\mathrm{Th},
\end{equation}
where $\mathrm{Th}=\ln(\frac{1-p}{p})$. Therefore, the overall support detection consists of multiple binary hypothesis testing instead of a single composite hypothesis testing. Replacing the $m$ binary hypothesis testing for determining the activity of each element renders a substantially lower computational complexity than using a composite hypothesis testing to search an activity vector $\qb$ over $2^m$ possible activity vectors.

\section{Amplitude Recovery}
Having detected the support vector, the amplitude of sparse vector is estimated. If we remove the inactive locations of sparse vector $\sbb$, we have an amplitude vector $\wb=[w_1,w_2,...,w_r]^T$, where $r$ is the number of active locations. If $I_a$ represents a set of indexes corresponding to active locations, we have the following relations:

\begin{equation}
\label{eq: reducedonebit}
\Bigg\{\begin{array}{cc}
                    y_1=\mathrm{sign}(\hb^T_1\wb+z_1), &  \\
                    y_2=\mathrm{sign}(\hb^T_2\wb+z_2), &  \\
                    ... &  \\
                    y_N=\mathrm{sign}(\hb^T_N\wb+z_N), & \qquad
                  \end{array}
\end{equation}
where $\hb_i=\ab_i(I_a)$ is the reduced measurement vector and $I_a=[0...1...0...1...1...]^T$ is the index of active
elements of $\sbb$. Now, we have the problem of estimating the non-sparse vector $\wb$ based on the sign measurements in (\ref{eq: reducedonebit}). This problem has been investigated in \cite{Zhu14} and an ML estimator has been introduced. The log-likelihood of measurements data can be obtained as \cite{Zhu14}
\begin{equation}
l(\yb;\wb)=\sum_{i=1}^N\log\Phi(\frac{y_i\hb_i^T\wb}{\sigma_z}).
\end{equation}
Maximizing the log-likelihood is equivalent to minimizing the negative log-likelihood. Consequently, the ML estimator is the solution of the following optimization problem:
\begin{equation}
\label{eq: p1}
\mathrm{P}_1: \quad \underset{\wb \in \mathbb{R}^{r}}{\text{minimize}}\quad -\sum_{i=1}^N\log\Phi(\frac{y_i\hb_i^T\wb}{\sqrt{||\wb||^2_2\sigma^2_e+\sigma^2_n}}),
\end{equation}
where the above optimization problem is non-convex and can not be solved by steepest-descent or Newton's method \cite{Zhu14}.
Moreover, in \cite{Zhu14}, it is proved that if we consider the unconstrained optimization problem as
\begin{equation}
\mathrm{P}_2: \quad \underset{\vb}{\text{minimize}}\quad -\sum_{i=1}^N\log\Phi(y_i\hb_i^T\vb),
\end{equation}
then the optimal point of problem $\mathrm{P}_1$ in (\ref{eq: p1}) exists if and only if the optimal point $\vb^{*}$ of problem $\mathrm{P}_2$ satisfies the constraint $||\vb^{*}||^2_2<\frac{1}{\sigma^2_e}$ \cite{Zhu14}. Therefore, we first solve the unconstrained optimization problem $\mathrm{P}_2$, and then check whether it satisfies the above-mentioned constraint to ensure whether the original ML estimation problem $\mathrm{P}_1$ in (\ref{eq: p1}) has an optimal point \cite{Zhu14} (For further details see \cite{Zhu14}).

\section{Simulation results}
\label{sec: Sim}
This section presents the simulation results. In the simulations, the unknown sparse vector $\sbb$ is drawn from a BG model with activity probability $p=0.1$ and $p=0.2$, and with variance of active samples $\sigma^2_r=1$. To ensure that the norm is finite, we normalized the sparse vector to have unit norm. The size of the sparse vector is assumed to be $m=200$. The sensing matrix elements are obtained from an standard Gaussian distribution with $a_{ij}\sim\mathcal{N}(0,1)$. The error matrix or perturbation matrix elements is considered to be $e_{ij}\sim\mathcal{N}(0,\sigma^2_e)$ with $\sigma_e=0.1$. The additive noise $\nb$ is regarded as Gaussian random variable with distribution $n_{i}\sim\mathcal{N}(0,\sigma^2_n)$, where $\sigma_n=0.1$.

We compare the proposed BHT-MLE method with the ML estimation method \cite{Zhu14} which is denoted as MLE. For the initialization of BHT-MLE and MLE, we use $\hat{\sbb}_0=\Ab^\dagger\yb$, where $\Ab^\dagger$ is the pseudo-inverse of matrix $\Ab$. To calculate the threshold $\mathrm{Th}=\ln(\frac{1-p}{p})$ in (\ref{eq: act}), we need to estimate $p$. Similar to \cite{ZayyB09}, we use $\hat{p}=\frac{\mathrm{Card}(\{s_i|s_i>\alpha\sqrt{\mathrm{var}(\sbb)}\})}{m}$, where $\mathrm{Card}$ is the cardinality operator. For initial iterations, we overestimate $p$ by choosing small value for $\alpha$ as $\alpha=0.5$. At final iterations, we choose $\alpha\approx3$. Extensive experimental studies demonstrate that $\alpha$ converges to its optimal value within 10 iterations. Hence, we use a linear increase of $\alpha$ as $\alpha^{(k)}=1.2\alpha^{(k-1)}$, where $k$ is the index of iteration. Similar to \cite{Zhu14}, we assume that the variances $\sigma^2_e$ and $\sigma^2_n$ are known in advance. For solving the unconstrained optimization problem $\mathrm{P}_2$, similar to \cite{Zhu14}, we use the MATLAB $\mathrm{fminunc}$ function.

\begin{figure}[tb]
\begin{center}
\includegraphics[width=9cm]{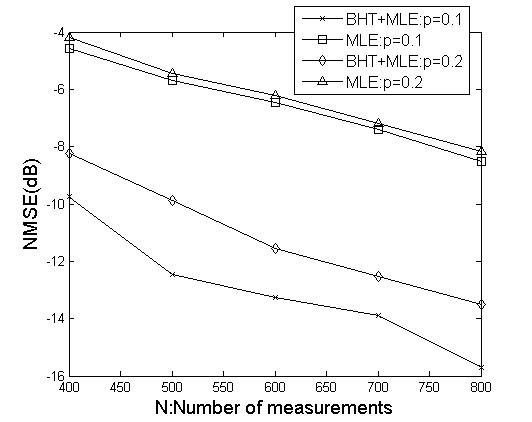}
\end{center}
\caption{Normalized mean square error of reconstructed sparse vector versus number of sign measurements.}
\label{fig3}
\end{figure}

We utilize the Normalized Mean Square Error (NMSE) as a performance metric, which is defined as
\begin{equation}
\mathrm{NMSE}\triangleq20\log_{10}(\frac{||\sbb-\hat{\sbb}||_2}{||\sbb||_2}),
\end{equation}
where $\hat{\sbb}$ is the estimate of true sparse vector $\sbb$.
 All the NMSEs are averaged over 100 Monte Carlo (MC) simulations. The number of binary measurements varies between 400 and 800. Figure~\ref{fig3} shows the NMSE performance versus the number of measurements for both BHT-MLE and MLE methods with $p=0.1$ and $p=0.2$. It is seen that the proposed BHT-MLE method, which utilizes Bayesian hypothesis testing, outperforms the conventional ML estimator (MLE) method by at least 5 dB gain because it exploits the activity information of the sparse vector provided by the first step of the algorithm. 
 To compare the complexity of the algorithms, we compute the average simulation time of MLE and BHT-MLE. For the case of $m=200$ and $N=400$, the average simulation times of MLE and BHT-MLE are 2.64 and 0.95 seconds, respectively. It shows that BHT-MLE is faster than MLE at least by a factor of two because it reduces the dimension of the optimization problem from $m=200$ to $r=\mathrm{round}(pN)=20$.

\section{Conclusion}
\label{sec: con}
We have proposed a new BHT-MLE algorithm for the noisy sparse signal reconstruction in one bit compressed sensing with sensing matrix perturbation framework. BHT-MLE algorithm consists of a Bayesian
hypothesis testing (BHT) for support detection and an ML estimator (MLE) for the amplitude estimation. In the support detection
step, the proposed BHT-MLE algorithm uses a sequence of binary hypothesis tests. Simulation results in a special case, verify that this BHT-based support detection approach
improves the sparse reconstruction accuracy by at least 5 dB gain and reduces the computational complexity by a factor of two.


\vskip5pt
\noindent H. Zayyani (\textit{Department of Electrical and Computer Engineering, Qom University of Technology, Qom, Iran})\\
\noindent E-mail: zayyani2009@gmail.com

\noindent M. Korki (\textit{Department of Telecommunications, Electrical, Robotics, and Biomedical Engineering, Swinburne University of Technology, Hawthorn, Australia})\\
\noindent F. Marvasti (\textit{Department of Electrical Engineering, Sharif University of Technology, Tehran, Iran})

\end{document}